\documentclass{article}

\PassOptionsToPackage{numbers, compress}{natbib}
\usepackage{nips_2017}

\usepackage[utf8]{inputenc} 
\usepackage[T1]{fontenc}    
\usepackage{hyperref}       
\usepackage{url}            
\usepackage{booktabs}       
\usepackage{amsfonts}       
\usepackage{nicefrac}       
\usepackage{microtype}      
\usepackage{color}
\usepackage{amsmath}
\usepackage{cases}
\usepackage{graphicx}
\usepackage{amsmath,epsfig}
\usepackage{color,float}
\usepackage{caption}
\def\eg{\emph{e.g.}}

\def\etal{\emph{et al.}}
\def\ie{\emph{i.e.}}

\title{A Self-Adaptive Proposal Model for Temporal Action Detection based on Reinforcement Learning}

\author{
$\bold{Jingjia\ Huang}^{1*}\quad \bold{Nannan\ Li}^{1}\thanks{means that the authors have equal contributions, and are listed in alphabetical order.}\quad \bold{Tao\ Zhang}^1\quad \bold{Ge\ Li}^1$ \\
${}^1 \rm{School\ of\ Electronic\ and\ Computer\ Engineering,\ Peking\ University\ Shenzhen\ Graduate\ School}$ \\
\texttt{jjhuang@pku.edu.cn, linn@pkusz.edu.cn, Z963895119@163.com, gli@pkusz.edu.cn} \\
}

\begin{document}

\maketitle

\begin{abstract}
Existing action detection algorithms usually generate action proposals through an extensive search over the video at multiple temporal scales, which brings about huge computational overhead and deviates from the human perception procedure. We argue that the process of detecting actions should be naturally one of observation and refinement: observe the current window and refine the span of attended window to cover true action regions. In this paper, we propose an active action proposal model that learns to find actions through continuously adjusting the temporal bounds in a self-adaptive way. The whole process can be deemed as an agent, which is firstly placed at a position in the video at random, adopts a sequence of transformations on the current attended region to discover actions according to a learned policy. We utilize reinforcement learning, especially the Deep Q-learning algorithm to learn the agent's decision policy. In addition, we use temporal pooling operation to extract more effective feature representation for the long temporal window, and design a regression network to adjust the position offsets between predicted results and the ground truth. Experiment results on THUMOS 2014 validate the effectiveness of the proposed approach, which can achieve competitive performance with current action detection algorithms via much fewer proposals.
\end{abstract}

\section{Introduction}
Temporal action detection requires not only to determine whether an action occurs in a video but also to locate the temporal extent of when it occurs, which is a challenging problem for real-life long untrimmed videos.
Most of modern approaches \citep{shou2016temporal, zhu2016efficient, xiong2017pursuit} usually solve the problem via a two-step pipeline: firstly generate a set of class independent action proposals, which are obtained via running a action/background classifier over a video at multiple temporal scales; then the proposals are classified by the pre-trained action detector, and post processing such as non-maximum suppression is applied. However, such extensive search for action localization is unsatisfying in terms of both accuracy and computational efficiency. Like the human detects the action through successively altering the span of attended region to narrow down the difference between the bounds of current window and that of true action region, the optimal algorithm should be the process of sequential, iterative observation and refinement consuming search steps as less as possible.

In this paper, we propose a class-specific action detection model that learns to continuously adjust the current region to cover the groundtruth more precisely in a self-adapted way. This is achieved by applying a sequence of transformations to a temporal window that is initially placed in the video at random and finally finds and covers action region as large as possible. The sequence of transformation is decided by an agent that analyzes the content of the current attended region and select the next best action according to a learned policy, which is trained via reinforcement learning based on Deep Q-Learning algorithm \citep{mnih2015human}. Different from existing approaches that locate the action following a fixed path, our method generates various search trajectories for different action instances, depending on the video scenarios, the starting search position and the sequences of actions adopted. As a result, the trained agent will locate a single instance of an action in about 15 steps, which means that the model only processes 15 successive regions of an image to explore an uncover video segment, thus it is of great computational efficiency to compare with sliding window based approaches.

Our model draws the inspiration from works that have used reinforcement learning to build active models for object localization in image \citep{caicedo2015active, jie2016tree, bellver2016hierarchical}. However, we can not handle the video in a top-down way that is proved to perform effectively for image object localization, as the duration of the video is usually too long (from hundreds to thousands frames). We start the search from a position randomly selected from the video, which will terminate until a instance of action has been found or the maximum transformation steps has been reached, and then a new search begins from the position away from current attended region. We incorporate temporal pooling operation with feature extraction process to better represent the long video segment and design a "\emph{jump}" action to avoid the agent trapping itself in the region where no action occurs. We conducted a comprehensive experimental evaluation in the challenging THUMOS'14 dataset \citep{THUMOS14}, and the results demonstrate that the proposed method can achieve competitive performance in terms of precision and recall via a small number of action proposals.

\section{Related work}
$\bold{Action\ Recognition.}$ This task has been attended for a few years, and a large amount of research work have been done \citep{laptev2003space, wang2013action, simonyan2014two, tran2015learning}. In early years, researchers often tackle the problem based on hand-crafted visual features \citep{laptev2003space, wang2013action}. Recently, impressed by the huge success of deep learning on image analysis task, some approaches have introduced deep models, especially Convolutional Neural Network (CNN), for better excavation the spatial-temporal information included in the video clip. Simonyan and Zisserman \citep{simonyan2014two} propose the two-stream network architecture with one branch processing RGB signal and the other one dealing with optical-flow signal. Tran $\etal$ \citep{tran2015learning} construct C3D model, which operates 3D convolution in spatio-temporal video volume directly and integrates appearance and motion cues for better feature representation. There have been also other efforts \citep{donahue2015long, yue2015beyond} that attempt to combine frame-level CNN feature representation and long-range temporal structure to cope with input videos of long duration. Up to now, deep learning based approaches have achieved state-of-the-art performances.

$\bold{Temporal\ Action\ Detection.}$
Different from action recognition where actions are included in a trimmed video clip and the aim is to predict the category, temporal action detection needs to not only classify the action but also give out temporal localization. Most existing approaches address the problem via sliding window strategy for candidates generation and focus on feature representation and classifier construction \citep{shou2016temporal, gaidon2013temporal, oneata2013action, yuan2016temporal}.
Shou $\etal$ \citep{shou2016temporal} utilize a multi-stage CNN detection network for action localization, where background windows are first filtered out by a binary action/background classifier based on C3D feature, then an action detection network incorporated both classification loss and temporal localization loss is trained for candidate refinement. By the limitation of 16-frames input of C3D model, they select 16 frames in uniform from the whole video, which is inferior to temporal pooling operation utilized in our approach. Gao $\etal$ \citep{gao2017turn} decompose the input video into short video units, and pool features extracted from a set of contiguous units for representation of long video clip, and meanwhile employ a coordinate regression network to refine the temporal action boundaries. Our approach also includes location regression, whose regression offsets are calculated via the relative deviation rather than the absolute value, thus it will facilitate the model to converge more efficiently. Unlike the works mentioned above, Yeung $\etal$ \citep{yeung2016end} propose an attention based model that predicts the action position through a few of glimpses, which is trained via reinforcement learning. The difference between their work and ours is that our approach locates the action through continuously adjusting the span of current window not predicting the bounds directly.

$\bold{Object\ Detection.}$
Most of recent approaches for object detection are built upon the paradigm of "\emph{proposal + classification}" \citep{girshick2014rich, ren2015faster}. Object proposals are usually either generated by methods relied on hand-crafted low-level visual cues, such as SelectiveSearch \citep{uijlings2013selective} and Edgebox \citep{zitnick2014edge}, or produced by fully convolutional network implemented on CNN features extracted from anchor boxes arranged uniformly on the image, such as Faster R-CNN \citep{ren2015faster}. However, generating too many proposals for a image with only one or two objects is unnecessary and computational inefficiency. Some works attempt to reduce the number of proposals with an active object detection strategy \citep{caicedo2015active, jie2016tree, mathe2016reinforcement}. Caicedo $\etal$ \citep{caicedo2015active} learns an optimal policy to locate one single object in the image via Deep Q-Learning, where it starts from the whole image in a top-down way and adaptively adjusts the window scale and position to focus on the true region. Jie $\etal$ \citep{jie2016tree} propose an effective tree-structured reinforcement learning approach, which learns to balance the exploration of uncovered new objects and the refinement of covered ones, and can localize multiple objects in a single run. Inspired by \citep{caicedo2015active, jie2016tree}, we design a reinforcement learning based approach for temporal action localization, which locates action instances within the long untrimmed video via the learned policy in a bottom-up way, and meanwhile utilizes a regression network to refine the predicted temporal window boundaries.

\begin{figure}[!t]
\centering
\includegraphics[height=0.25\textwidth,width=1.0\textwidth]{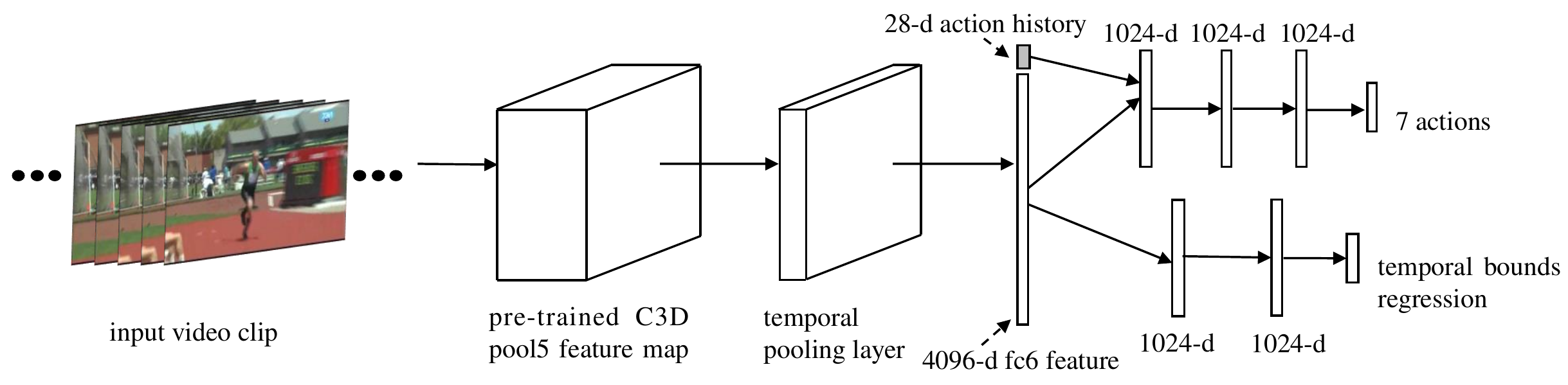}
\caption{The framework of our proposed action proposal model based on Deep Q-learning, which incorporates a regression network for better action localization.}
\label{frameworknip}
\end{figure}

\section{Self-Adaptive Action Proposal Model}
In this section, we present our action-proposal generation model, which is self-adapted and will gradually adjust its predicted results according to the content of attended window and the history of executed actions to cover the true action region as accurate as possible in a few steps. We cast the problem of temporal action localization as a Markov Decision Process (MDP), in which the agent interacts with the environment and makes a sequence of decisions to achieve the settled goal. In our formulation, the environment is the input video clip, in which the agent has an observation of the current video segment, called temporal window, and restructures the position or span of the window, to achieve the goal of locating the action precisely. The agent receives positive or negative rewards after each decision made during the train phase to learn an effective policy.
Besides, we construct a regression network to refine the final detection results to promote the accuracy of localization. The framework of our proposal generation model is illustrated in Fig. \ref{frameworknip}. In the following subsections, the set of actions $A$, the set of states $S$, and the reward function $R(s,a)$ of MDP and the regression network are discussed in detail. To avoid confusion, the action performed by the actor in the video is called motion in this section, and in other sections the meaning of action is determined by the context.

\subsection{MDP Formulation}
\begin{figure}[!t]
\centering
\includegraphics[width=1.0\textwidth]{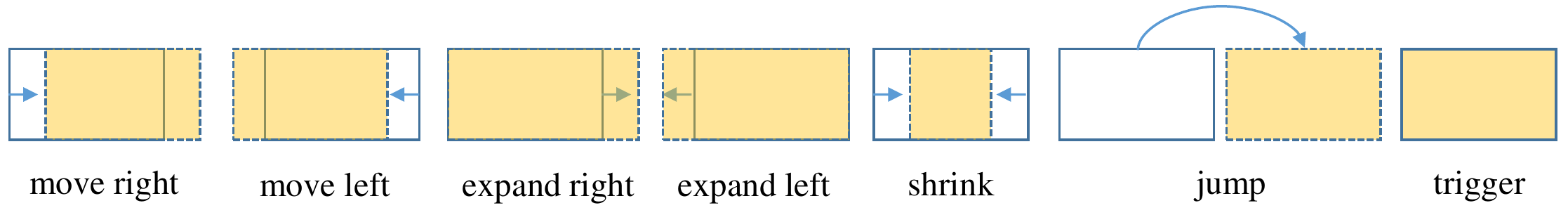}
\caption{Illustrations of the actions adopted by the agent for motion search in our experiment. Yellow windows with dash line represent next windows after taking the corresponding action.}
\label{actions}
\end{figure}

$\bold{Actions}.$ The set of actions $A$ can be divided into two categories: one group for transformation on temporal window, such as "$\emph{move left}$", "$\emph{move right}$", "$\emph{expand left}$", and the remaining one for terminating the search, "$\emph{trigger}$", as shown in Fig. \ref{actions}. The transformation group includes regular actions that comprises of translation and scale, and one irregular action. The regular actions vary the current window in terms of position or time span around the attended region, such as "$\emph{move left}$", "$\emph{expand left}$" or "$\emph{shrink}$", which are adopted by the agent to increase the intersection with the groundtruth that has overlaps with the current window. The irregular action, namely "$\emph{jump}$", translates the window to a new position away from the current site to avoid that the agent traps itself round the present location when there is no motion occurring nearby. The change caused by any regular actions at each time to the window equals to a value in proportion to the current window size. For instance, supposing that current window is denoted as $[x_l, x_r]$, where $x_l$ and $x_r$ stand for the left and right boundary respectively. The action "$\emph{move left}$" translates the window to a new site of $[x_{l'}, x_{r'}]$ with $x_{l}-x_{l'} = x_{r'}-x_r = \alpha \ast (x_r-x_l)$, while for action "$\emph{expand left}$" scales the window with the change of $x_l-x_{l'}=\alpha \ast (x_r-x_l)$ and $x_{r'} = x_r$. Here, $\alpha \in [0,1]$ is a parameter that can give a trade-off between search speed and localization accuracy. In this paper, we set $\alpha = 0.2$. The action "$\emph{jump}$" selects a new window randomly from the left or right side, which has the same size with the current window, being a distance away from the present site. The regular actions make the agent gradually adjust its position to cover the motion more accurately when it has found one; while the action "$\emph{jump}$" let the agent explore unknown region that may contain the motion in a discontinuous and efficient way. The action "$\emph{trigger}$" is employed by the agent whenever it considers that a motion has been localized by the current window, and stops the sequence of current search, and restarts a new search for the next motion with an initial window position away from current site.

$\bold{State}.$ The state of MDP is the concatenation of two components: the presentation of current window and the history of taken actions. To describe the motion within current window generally, the feature extracted from the C3D CNN model \citep{tran2015learning}, which is pretrained on Sports-1M and finetuned on UCF-101, is utilized as the presentation. Here, we choose the feature vector from $\emph{fc6}$ layer (4096 dimension) in our problem, consideing its good abstract representation for the semantic information about the motion. The original C3D model can only accept 16 frames as input, however, the duration of temporal window is always far more than that number. To tackle with the problem, we design two different solutions: i.) uniformly select 16 frames from the whole duration ; ii.) fed all the frames into the C3D model and add a additional pooling layer (average pooling for our problem) between the "\emph{pool5}" layer and the "\emph{fc6}" layer, which condenses the dimension of extracted feature vector from "\emph{pool5}" to the value specified by the C3D model. The history of the taken actions is a binary vector that tells which action has been adopted by the agent in the past. Each action in the history is represented by a 7-dimension binary vector where all the values are zero except the one corresponding to the taken action. In the experiment, we totally record 10 past actions as the history. The history of taken actions informs the agent the search path that has been passed through and the regions already attended, so as to stabilize search trajectories that might get stuck in repetitive cycles.

$\bold{Reward\ Fuction}.$ The reward function $R(s,a)$ provides a feedback to the agent when it performs the action $a$ at the current state $s$, which awards the agent for actions that will bring about the improvement of motion localization accuracy while gives the punishment for actions that leads to the decline of the accuracy. The quality of motion localization is evaluated via the simple yet indicative measurement, Intersection over Union (IoU) between current attended temporal window and the groundtruth. Supposing that $w$ stands for the current window and $g$ represents the groundtruth region of motion, then the IoU between $w$ and $g$ is defined as
IoU$(w,g)$ = span($w \cap g$) / span($w\cup g$). The reward function is proportional to the difference between IoUs of two successive states $s$ and $s'$, where the agent moves to state $s'$ from $s$ by executing the action $a$. Specially, it is formulated as following:

\begin{equation}\label{reward}
r(s,a) = \max\limits_{1\leq i\leq n} {\rm sign}({\rm IoU}(w',g_i)-{\rm IoU}(w,g_i)),
\end{equation}

where $w'$ and $w$ are attended windows corresponding to state $s'$ and $s$ respectively, $n$ is the number of groundtruths within the input video. The reward function returns +1 or -1. Equation \ref{reward} indicates that the agent receives the reward +1 if the new window $w'$ has more overlap with any of the groundtruth than the previous window $w$, while the reward -1 otherwise. Such binary reward value makes the agent clearly realize that at present state, which action drives the attended window towards the groundtruth, and thus accelerates the convergence rate of the model during training phase. In addition, such reward-function scheme facilitates better localization towards motion regions especially for the video with multiple motion instances, as there is no limitation on which motion should be focused on at each state. The "\emph{trigger}" action has a different reward function scheme, as it leads to the termination of search and there is no next state. The reward of "\emph{trigger}" is determined by a piecewise function of IoU threshold, which can be presented as following:

\begin{equation} \label{end}
r_e(s)=
\begin{cases}
+\eta& \text{if\ IoU$(w,g)\ \geq\ \tau$}\\
-\eta& \text{otherwise}.
\end{cases}
\end{equation}

 In equation \ref{end}, $e$ represents the "\emph{trigger}" action, $\eta$ is the reward value and chosen as 3 in our experiment, $\tau$ is the IoU threshold, which controls the tradeoff between the localization accuracy and computational overhead. The large $\tau$ will encourage the agent to locate the motion more precisely, however it consumes more action steps to complete the search. In training phase, we do not stop the search process when the agent correctly performs the action "\emph{trigger}" for the first time, and let it continuously explore uncover regions. Therefore, our model recognizes many termination states that have IoU with groundtruth more than $\tau$. We utilize $\tau = 0.5$ for our problem, and find that larger $\tau$, such as 0.6 or 0.7, gives rise to negligible promotion on recall value, which is validated by the experiments.

$\bold{Deep\ Q}$-$\bold{learning.}$ The goal of the agent is to maximize the sum of discounted rewards that are received through continuously transforming the current attended window during a sequence of interactions with the environment (an episode). In other words, the agent needs to learn a policy $\pi(s)$ that specifies an optimal action at current state $s$ in the view of maximizing the long-term benefit.
Due to the lack of state transition probability and the model free environment, we utilize reinforcement learning, specially Deep Q-learning, to estimate the optimal value for each state-action pair. In this paper, we follow the deep Q-learning framework proposed by Mnih $\etal$ \citep{mnih2015human} that estimates the action-value function via a neural network. The architecture of our Deep Q-Network (DQN) is illustrated as the up branch of Fig. \ref{frameworknip}. Similar to \citep{caicedo2015active, jie2016tree}, the C3D CNN model is just used for feature extraction, and we do not train the whole pipeline for the full feature hierarchy learning, due to the good generalization of CNN model pretrained on large dataset and short of sufficient motion detection data for jointly training both two networks. During training phase, the agent operates multiple episodes with randomly initialized positions for each video clip. We train separate DQN for each motion category and follow the $\epsilon$-greedy policy. Specially, the agent randomly selects an action from the whole action set with probability $\epsilon$ at current state, while greedily chooses the optimal action according to the learned policy with probability 1-$\epsilon$. During the whole training epochs, $\epsilon$ is annealed linearly from 1.0 to 0.1, which gradually shifts from exploration to exploitation. Following \citep{mnih2015human}, we also incorporate the replay-memory scheme to collect various transition experiences from the past episodes, from which each record may be repeatedly used for  model updates, in favor of breaking short-term correlations between states. A minibatch ($\eg$ 200 records) is randomly sampled from replay-memory as training samples to update the model at each time.

\subsection{Regression Network}
Inspired by Fast R-CNN \citep{girshick2014rich} where a regression network is incorporated to revise the position deviation between the predicted result and the groundtruth, we also introduce a regression model to refine the motion proposals. As shown in the down branch of Fig. \ref{frameworknip}, the regression channel accepts 4096-dimension feature vector as input and gives out two coordinate offsets on both starting and end moment. Unlike spatial bounding box regression, in which coordinate scaling is needed due to various camera-projection perspectives, we directly utilize original temporal coordinate ($\ie$ frame number) for offsets calculation leveraging the advantage of  unified frame rate among video clips in our experiment. The regression biases are represented as the ratio of position deviation relative to the predicted span, which are defined as following:

\begin{equation}\label{regression}
o_s = (s_p-s_g)/(e_p-s_p), \quad  o_e = (e_p-e_g)/(e_p-s_p),
\end{equation}

where $s_p$ and $e_p$ are frame indexes for predicted starting and end moment, while $s_g$ and $e_g$ are frame indexes for the matched groundtruth. The loss for temporal coordinate regression, $L_{reg}$, is defined as following:

\begin{equation} \label{lossreg}
L_{reg} = \frac{1}{N_{end}} \sum^{N_{end}}_{i=1}(|o_{s,i}| + |o_{e,i}|),
\end{equation}

where $N_{end}$ is the number of actions that correctly perform "$trigger$" in a minibatch. In other words, we only regress the position of temporal window whose IoU with groundtruth is more than 0.5. We utilize $L_{1}$ norm to make the loss be insensitive to outliers.

\section{Experimental Results}
\label{experiments}
We evaluate the performance of our model on the dataset THUMOS'14. Followed the standard evaluation protocol, our method achieves a competitive recall compared to the state-of-the-art results and outperforms the existing methods by a large margin for action detection task.
\subsection{Implementation Details}
\textbf{Datasets.} We validate the quality of our methods on labeled untrimmed videos from the challenging THUMOS'14, which contains over 20 hours of video from 20 sport action categories. The dataset comprises 413 videos with 200 for validation and 213 for test.
 We train our model on validation set and report results on test set.\\
\textbf{Training Details.} Our model is implemented on Torch 7 \cite{collobert2011torch7}. We train category specific model for each action and keep the same parameters settings. In pre-process stage, we downsample videos to extract a more compact C3D descriptor. The replay memory buffer size is set as 2000, while the minibatch size is 200. The learning rate for DQN is 1e-3 with a decay rate of 5e-5, while the learning rate for regression network is 1e-4 with a decay rate of 9e-5. Dropout is applied with a ratio of 0.2. To accelerate training, we force the agent to take a "\emph{jump}" action if the IoU for current window is zero, which will drive the window to the region around a groundtruth.  \\
\textbf{Testing Details.} During test phase, the agent starts its search from the beginning, and take actions to adjust itself position according to the attended region. We set a maximum action steps for the agent as 15. The agent will restart its search from the back bound of current window, if it takes a "\emph{trigger}" action or finishes maximum action steps. Note that different from training, the agent consistently takes a leap forward (two times farther than move "\emph{left/right}") when it adopts a "\emph{jump}" action. We choose the windows, where the agent takes "\emph{trigger}" action, as proposals and utilize the pre-trained TSN \cite{Wang2016Temporal} as our classifier.

\subsection{ Temporal Proposals Evaluation}
\begin{figure}[t]
   \captionsetup{font={small}}
\begin{minipage}[b]{.5\linewidth}
  \centering
\centerline{\epsfig{figure=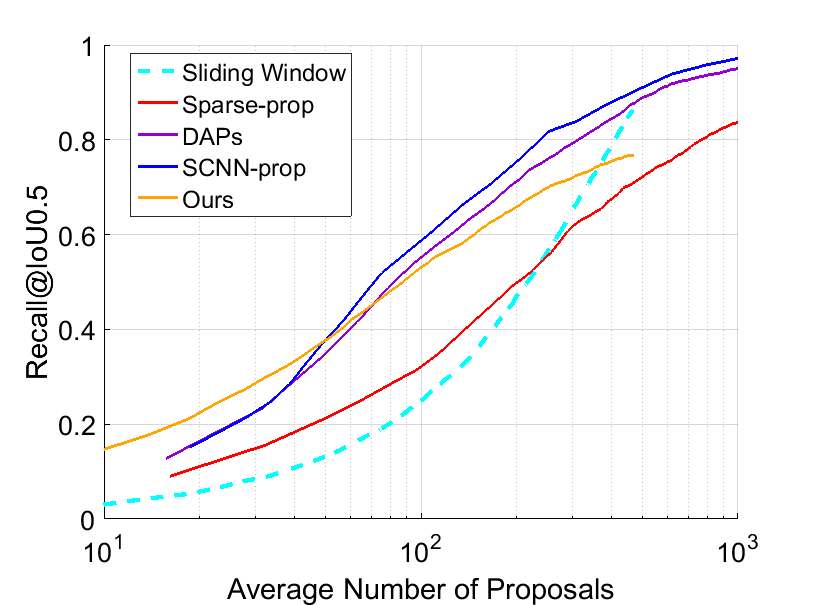,width=6.5cm}}
  \vspace{0.2cm}
  \begin{small}
  \centerline{(a)}\medskip
  \end{small}
\end{minipage}
\hfill
\begin{minipage}[b]{.5\linewidth}
  \centering
\centerline{\epsfig{figure=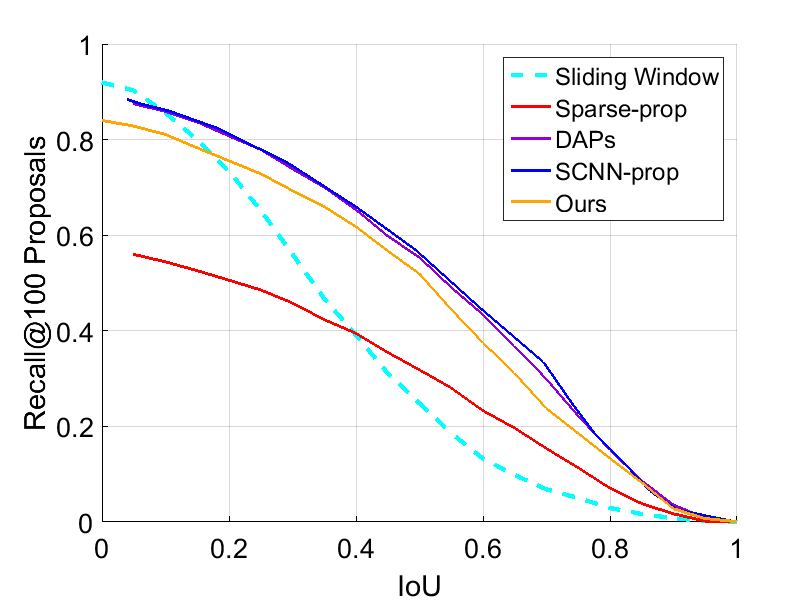,width=6.5cm}}
  \vspace{0.2cm}
  \begin{small}
  \centerline{(b)}\medskip
  \end{small}
\end{minipage}
\caption{Evaluation results of recall performance on THUMOS'14. S-CNN and DAPs are state-of-the-art methods while \emph{Sliding Window} in dash line is the baseline. We use the codes provided by \citep{Escorcia2016DAPs} to calculate recalls}
\label{fig:Recall}
\end{figure}

All the regions attended by the agent can be understood as temporal proposal candidates. Our methods run for about 400 steps with around 50 triggers averaged for each video. Fig.\ref{fig:example} is an instance of the detection process of DQN. For each attended region, we score them with the Q-value predicted by the model, and add a large bonus only to "\emph{trigger}" regions in order to give them higher priority when ranking the proposals. To assess the recall performance of our method, we use the metrics from \cite{Escorcia2016DAPs}:\\
\textbf{Recall vs. Average Number of Proposal}: average recall over all categories at IoU 0.5 is calculated as a function of average number of proposal. The best proposal approach is expected to achieve a higher recall with less proposals.\\
\textbf{Recall vs. IoU}: for a fixed number of proposals, recall is calculated at IoU between 0.05 to 1. To measure the localization quality of the top ranked proposals that are of most important for further recognition task, we fix the number of proposals to 100.\\
We compare our method with DAPs \citep{Escorcia2016DAPs}, SCNN-prop \citep{shou2016temporal}, Sparse-prop \cite{Heilbron2016Fast} and \emph{sliding window}. SCNN-prop and DAPs are the state-of-the-art methods while \emph{sliding window} is the baseline. For DAPs, SCNN-prop and Sparse-prop, we plot the curves using the proposal results provided by the authors. \emph{sliding window} generates the proposals including all siding windows of lengths from 16 to 512 with 50\% overlapping, and each window is scored with a random value. As shown in Fig.\ref{fig:Recall}.(a), our method achieves a better performance than the state-of-the-art methods in the early state of recall, and we have a competitive recall performance for the top 100 proposals according to Fig.\ref{fig:Recall}.(b). Notice that, the recall growth of our method slows down after about 70 proposals. It is because that our DQN agent tries to figure out the ground truth as fast as possible, and tends to stop exploration when it considers that an action region with IoU more than 0.5 has been found. Therefore, except the "\emph{trigger}" segments, other proposals are intermediate results during the exploring process, which are unreliable on most occasions.

\subsection{Temporal Action Detection Analysis}
\begin{figure}[t]
\centering
\includegraphics[width=14cm]{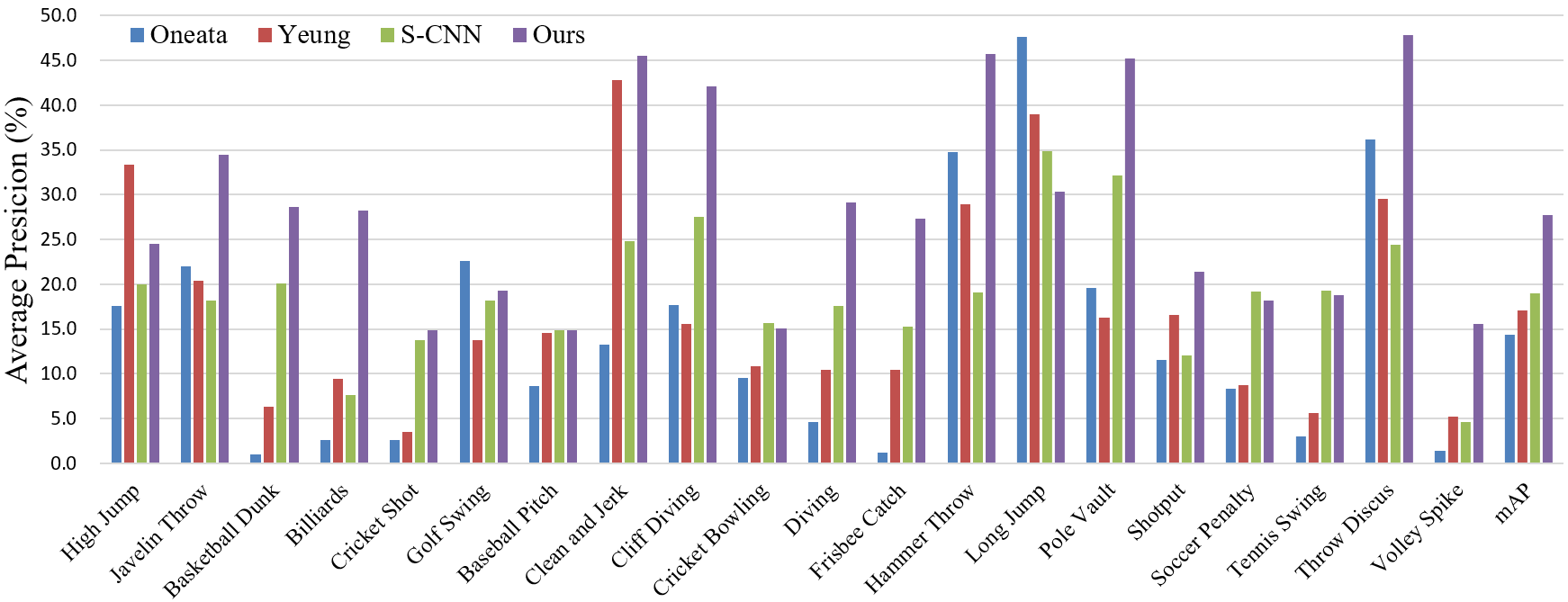}
\caption{\footnotesize{Histograms of average precision  for each categories on THUMOS'14. The results are calculated with the official toolkit. The mAP(\%) for Oneata \emph{et al.} \cite{oneata2013action}, Yeung \emph{et al.} \cite{yeung2016end}, S-CNN \citep{shou2016temporal} and Ours are 14.4, 17.1, 19.0 and 27.7 respectively.}}
\label{fig:mAP}
\end{figure}

 Following the convention \cite{THUMOS14}, we evaluate the performance of our methods on the temporal localization task with mean Average Precision(mAP) score at 50\% IoU. In the experiments, we take "\emph{trigger}" windows as proposals and classify them with a pre-trained TSN. Our methods are compared with other state-of-the-art methods in the literature, including S-CNN \citep{shou2016temporal}, Oneata \emph{et al.} \cite{oneata2013action} and Yeung \emph{et al.} \cite{yeung2016end}. As shown in Fig.\ref{fig:mAP}, our method outperforms the state-of-the-art approaches on THUMOS'14 by a large margin of 8.4\%.

To further analyze the contributions of different model components, namely temporal pooling and coordinate regression, for action detection task, we implement ablation studies.  We construct three models, which are described as follows:\\
$\bullet$ \textbf{Ours}: The integrated model with architecture shown in Fig. \ref{frameworknip}, where DQN agent generates proposals with the features processed through temporal pooling layer and finetunes the proposals with regression network. Average number of "\emph{trigger}" proposals is 67 per video.\\
$\bullet$ \textbf{Ours-POOLING}: The model without temporal pooling layer, uniformly samples video frames from input video segment to extract C3D features
and finetunes the proposals with regression network. Average number of "\emph{trigger}" proposals is 50 per video.\\
$\bullet$ \textbf{Ours-POOLING-RGN}: The basic model without both temporal pooling layer and regression network, uniformly samples video frames and only utilizes DQN for proposal generation.
 Average number of "\emph{trigger}" proposals is 50 per video.\\
For each model, we evaluate the proposals and overall action localization performance. First of all, we use \textbf{Recall vs. Average Number of Proposal} at IoU=0.5 to evaluate the proposal performance that is shown in Fig. 5. Then we present the quantitative detection results of the models in Table 1 that are reported by mAP scores at 50\% IoU. The mAPs are calculated with a fixed number of average proposals ($i.e.$ 50) that is equal to the number of average "\emph{trigger}" proposals for \textbf{Ours-POOLING} and \textbf{Ours-POOLING-RGN}. The last line in Table 1 reports the mAP calculated with the total "\emph{trigger}" proposals for \textbf{Ours} where average number is 67. Interestingly, it seems that \textbf{Ours-POOLING} has the superior recall performance than \textbf{Ours}, as shown in Fig. 5, however, \textbf{Ours} outperforms the other models by a large margin on overall detection performance. As pointed out by \citep{Escorcia2016DAPs}, we consider that this inconsistent result claims that \textbf{Ours} produces proposals with a small number of hard negatives, which allows the activity classifier to keep the number of false positive low. Besides, the results also illustrate that localization regression is of benefit to the detection task without exception. We also compare our proposal models with exiting related works, and our model achieves the best performance for action detection task, as shown in Table 1.

\begin{figure}
\begin{minipage}[b]{.5\textwidth}
\centering
\centerline{\epsfig{figure=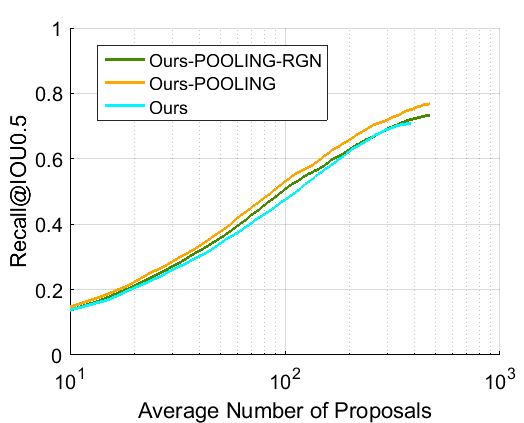,width=5cm}}
\caption{\small{Recall Evaluation for ablation study: \qquad \qquad \\ Recall vs. Average Number of Proposal at IoU=0.5}}
\end{minipage}
\label{fig:Ablation}
\makeatletter\def\@captype{table}\makeatother
\begin{minipage}[b]{.5\textwidth}
\centering
\tiny
\begin{tabular}{c|c}

  \hline
  Model@Proposal Number & mAP(\%) \\
  \hline
  SCNN \citep{shou2016temporal}@NA  & 19.0 \\
  \hline
  Yeung \emph{et al.} \cite{yeung2016end}@NA  & 17.1 \\
  \hline
  Oneata \emph{et al.} \cite{oneata2013action}@NA & 14.4\\
  \hline

  Ours-POOLING-RGN@50 & 22.3 \\
  \hline
  Ours-POOLING@50 &  24.6 \\
  \hline
  Ours@50  & \textbf{26.4} \\
  \hline
  Ours@67  & \textbf{27.7} \\
      \hline
\end{tabular}
\label{table:Ablation}
\caption{\footnotesize {Temporal-action detection results evaluation for various proposal models: mAP calculated with a fixed number of average proposals on THUMOS'14. @NA means the proposal number is not specified for the methods.}}
\end{minipage}
\end{figure}

\begin{figure}[t]
\centering
\includegraphics[width=14.1cm]{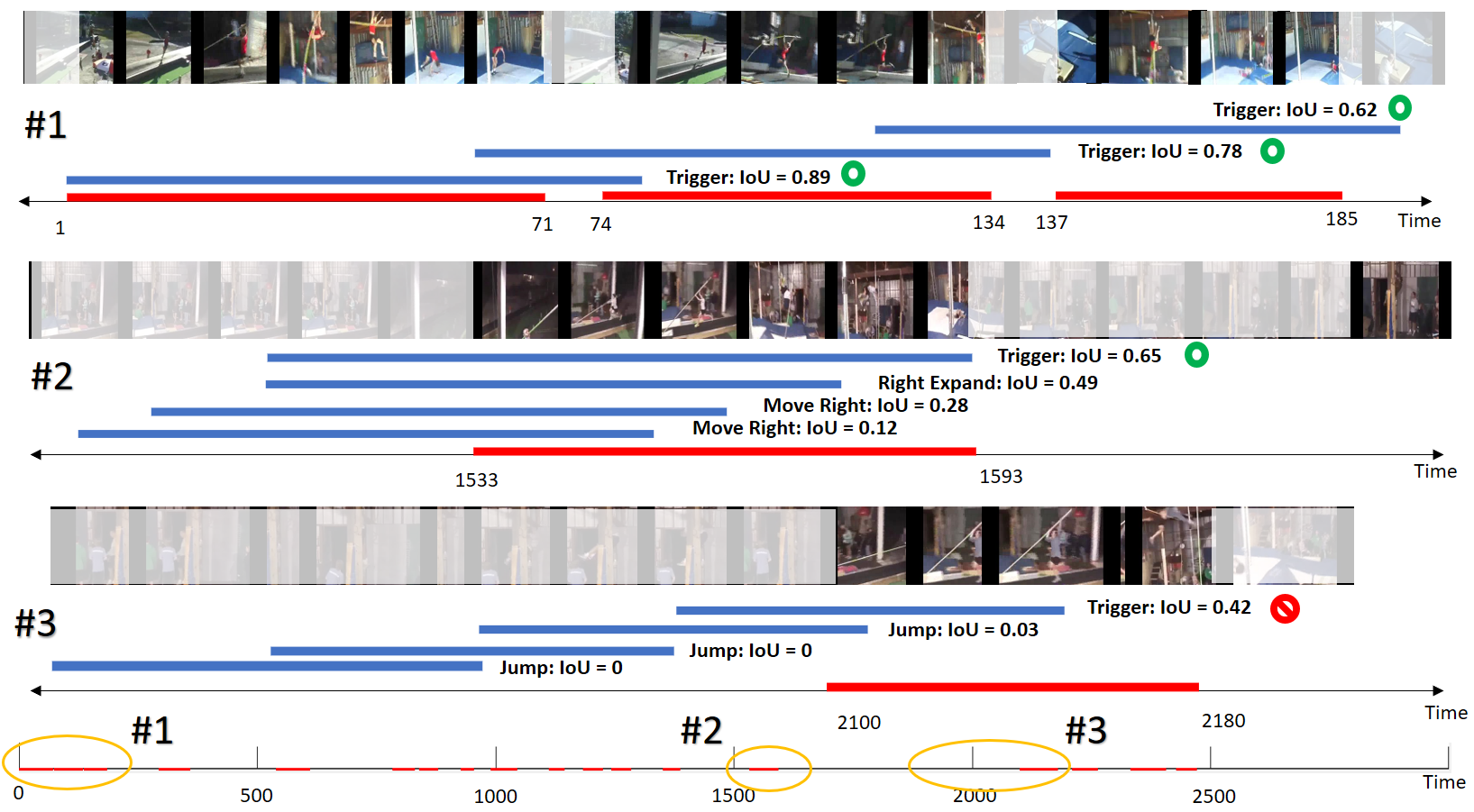}
\caption{\footnotesize{An instance of how DQN agent takes actions to generate proposals. The examples are sampled from the action \emph{PoleVaul} of THUMOS'14. The last row is the time line, where the red lines correspond to the ground truth. The top 3 rows are the running details corresponding to action instances $\#1$, $\#2$ and $\#3$. The blue lines are the agent's search histories. A green circle indicates that it is a right "\emph{trigger}" decision while the red one indicates a wrong one.}}
\label{fig:example}
\end{figure}

\subsection{Run-time Performance}
The run-time property of our method is dependent on the DQN's performance. For a well trained DQN agent, it will concentrate on the ground truth in a couple of steps once it perceives the action segment. Meanwhile, it can also accelerate the exploring process over the video with "jump" action. Besides, the selection of scalar $\alpha$ is also an important factor that will influence the run-time performance. A large $\alpha$ will make the agent take a brief glance over the video in most of the case, but will also result in coarse proposals. As a trade off,we set the $\alpha$ = 0.2 during the training and testing phase. On Tesla K80 platform, the average run-time of our model over all testing videos in THUMOS'14 is 50.4 FPS, including the online C3D extraction.

\section{Conclusion}
In this paper, we have introduced an active action proposal model that learns to adaptively adjust the span of attended current window to cover the true action regions in a few steps. We build our model based on deep reinforcement learning and lean an optimal policy to direct the agent to act. In order to precisely locate the action, we design a regression network to revise the offsets between predicted bound results and the groundtruth. Experiment results on THUMOS 14 dataset validate that the proposed approach can achieve comparable performance with most of modern action-detection methods with much fewer action proposals.

\small
\bibliographystyle{unsrtnat}
\setlength{\bibsep}{0.5ex}
\bibliography{nips}

\begin{thebibliography}{28}
\providecommand{\natexlab}[1]{#1}
\providecommand{\url}[1]{\texttt{#1}}
\expandafter\ifx\csname urlstyle\endcsname\relax
  \providecommand{\doi}[1]{doi: #1}\else
  \providecommand{\doi}{doi: \begingroup \urlstyle{rm}\Url}\fi

\bibitem[Shou et~al.(2016)Shou, Wang, and Chang]{shou2016temporal}
Zheng Shou, Dongang Wang, and Shih-Fu Chang.
\newblock Temporal action localization in untrimmed videos via multi-stage
  cnns.
\newblock In \emph{Proceedings of the IEEE Conference on Computer Vision and
  Pattern Recognition}, pages 1049--1058, 2016.

\bibitem[Zhu and Newsam(2016)]{zhu2016efficient}
Yi~Zhu and Shawn Newsam.
\newblock Efficient action detection in untrimmed videos via multi-task
  learning.
\newblock \emph{arXiv preprint arXiv:1612.07403}, 2016.

\bibitem[Xiong et~al.(2017)Xiong, Zhao, Wang, Lin, and Tang]{xiong2017pursuit}
Yuanjun Xiong, Yue Zhao, Limin Wang, Dahua Lin, and Xiaoou Tang.
\newblock A pursuit of temporal accuracy in general activity detection.
\newblock \emph{arXiv preprint arXiv:1703.02716}, 2017.

\bibitem[Mnih et~al.(2015)Mnih, Kavukcuoglu, Silver, Rusu, Veness, Bellemare,
  Graves, Riedmiller, Fidjeland, Ostrovski, et~al.]{mnih2015human}
Volodymyr Mnih, Koray Kavukcuoglu, David Silver, Andrei~A Rusu, Joel Veness,
  Marc~G Bellemare, Alex Graves, Martin Riedmiller, Andreas~K Fidjeland, Georg
  Ostrovski, et~al.
\newblock Human-level control through deep reinforcement learning.
\newblock \emph{Nature}, 518\penalty0 (7540):\penalty0 529--533, 2015.

\bibitem[Caicedo and Lazebnik(2015)]{caicedo2015active}
Juan~C Caicedo and Svetlana Lazebnik.
\newblock Active object localization with deep reinforcement learning.
\newblock In \emph{Proceedings of the IEEE International Conference on Computer
  Vision}, pages 2488--2496, 2015.

\bibitem[Jie et~al.(2016)Jie, Liang, Feng, Jin, Lu, and Yan]{jie2016tree}
Zequn Jie, Xiaodan Liang, Jiashi Feng, Xiaojie Jin, Wen Lu, and Shuicheng Yan.
\newblock Tree-structured reinforcement learning for sequential object
  localization.
\newblock In \emph{Advances in Neural Information Processing Systems}, pages
  127--135, 2016.

\bibitem[Bellver et~al.(2016)Bellver, Gir{\'o}-i Nieto, Marqu{\'e}s, and
  Torres]{bellver2016hierarchical}
Miriam Bellver, Xavier Gir{\'o}-i Nieto, Ferran Marqu{\'e}s, and Jordi Torres.
\newblock Hierarchical object detection with deep reinforcement learning.
\newblock \emph{arXiv preprint arXiv:1611.03718}, 2016.

\bibitem[Jiang et~al.(2014)Jiang, Liu, Roshan~Zamir, Toderici, Laptev, Shah,
  and Sukthankar]{THUMOS14}
Y.-G. Jiang, J.~Liu, A.~Roshan~Zamir, G.~Toderici, I.~Laptev, M.~Shah, and
  R.~Sukthankar.
\newblock {THUMOS} challenge: Action recognition with a large number of
  classes.
\newblock \url{http://crcv.ucf.edu/THUMOS14/}, 2014.

\bibitem[Laptev and Lindeberg(2003)]{laptev2003space}
Ivan Laptev and Tony Lindeberg.
\newblock Space-time interest points.
\newblock In \emph{9th International Conference on Computer Vision, Nice,
  France}, pages 432--439. IEEE conference proceedings, 2003.

\bibitem[Wang and Schmid(2013)]{wang2013action}
Heng Wang and Cordelia Schmid.
\newblock Action recognition with improved trajectories.
\newblock In \emph{Proceedings of the IEEE International Conference on Computer
  Vision}, pages 3551--3558, 2013.

\bibitem[Simonyan and Zisserman(2014)]{simonyan2014two}
Karen Simonyan and Andrew Zisserman.
\newblock Two-stream convolutional networks for action recognition in videos.
\newblock In \emph{Advances in neural information processing systems}, pages
  568--576, 2014.

\bibitem[Tran et~al.(2015)Tran, Bourdev, Fergus, Torresani, and
  Paluri]{tran2015learning}
Du~Tran, Lubomir Bourdev, Rob Fergus, Lorenzo Torresani, and Manohar Paluri.
\newblock Learning spatiotemporal features with 3d convolutional networks.
\newblock In \emph{Proceedings of the IEEE International Conference on Computer
  Vision}, pages 4489--4497, 2015.

\bibitem[Donahue et~al.(2015)Donahue, Anne~Hendricks, Guadarrama, Rohrbach,
  Venugopalan, Saenko, and Darrell]{donahue2015long}
Jeffrey Donahue, Lisa Anne~Hendricks, Sergio Guadarrama, Marcus Rohrbach,
  Subhashini Venugopalan, Kate Saenko, and Trevor Darrell.
\newblock Long-term recurrent convolutional networks for visual recognition and
  description.
\newblock In \emph{Proceedings of the IEEE conference on computer vision and
  pattern recognition}, pages 2625--2634, 2015.

\bibitem[Yue-Hei~Ng et~al.(2015)Yue-Hei~Ng, Hausknecht, Vijayanarasimhan,
  Vinyals, Monga, and Toderici]{yue2015beyond}
Joe Yue-Hei~Ng, Matthew Hausknecht, Sudheendra Vijayanarasimhan, Oriol Vinyals,
  Rajat Monga, and George Toderici.
\newblock Beyond short snippets: Deep networks for video classification.
\newblock In \emph{Proceedings of the IEEE conference on computer vision and
  pattern recognition}, pages 4694--4702, 2015.

\bibitem[Gaidon et~al.(2013)Gaidon, Harchaoui, and Schmid]{gaidon2013temporal}
Adrien Gaidon, Zaid Harchaoui, and Cordelia Schmid.
\newblock Temporal localization of actions with actoms.
\newblock \emph{IEEE transactions on pattern analysis and machine
  intelligence}, 35\penalty0 (11):\penalty0 2782--2795, 2013.

\bibitem[Oneata et~al.(2013)Oneata, Verbeek, and Schmid]{oneata2013action}
Dan Oneata, Jakob Verbeek, and Cordelia Schmid.
\newblock Action and event recognition with fisher vectors on a compact feature
  set.
\newblock In \emph{Proceedings of the IEEE International Conference on Computer
  Vision}, pages 1817--1824, 2013.

\bibitem[Yuan et~al.(2016)Yuan, Ni, Yang, and Kassim]{yuan2016temporal}
Jun Yuan, Bingbing Ni, Xiaokang Yang, and Ashraf~A Kassim.
\newblock Temporal action localization with pyramid of score distribution
  features.
\newblock In \emph{Proceedings of the IEEE Conference on Computer Vision and
  Pattern Recognition}, pages 3093--3102, 2016.

\bibitem[Gao et~al.(2017)Gao, Yang, Sun, Chen, and Nevatia]{gao2017turn}
Jiyang Gao, Zhenheng Yang, Chen Sun, Kan Chen, and Ram Nevatia.
\newblock Turn tap: Temporal unit regression network for temporal action
  proposals.
\newblock \emph{arXiv preprint arXiv:1703.06189}, 2017.

\bibitem[Yeung et~al.(2016)Yeung, Russakovsky, Mori, and Fei-Fei]{yeung2016end}
Serena Yeung, Olga Russakovsky, Greg Mori, and Li~Fei-Fei.
\newblock End-to-end learning of action detection from frame glimpses in
  videos.
\newblock In \emph{Proceedings of the IEEE Conference on Computer Vision and
  Pattern Recognition}, pages 2678--2687, 2016.

\bibitem[Girshick et~al.(2014)Girshick, Donahue, Darrell, and
  Malik]{girshick2014rich}
Ross Girshick, Jeff Donahue, Trevor Darrell, and Jitendra Malik.
\newblock Rich feature hierarchies for accurate object detection and semantic
  segmentation.
\newblock In \emph{Proceedings of the IEEE conference on computer vision and
  pattern recognition}, pages 580--587, 2014.

\bibitem[Ren et~al.(2015)Ren, He, Girshick, and Sun]{ren2015faster}
Shaoqing Ren, Kaiming He, Ross Girshick, and Jian Sun.
\newblock Faster r-cnn: Towards real-time object detection with region proposal
  networks.
\newblock In \emph{Advances in neural information processing systems}, pages
  91--99, 2015.

\bibitem[Uijlings et~al.(2013)Uijlings, Van De~Sande, Gevers, and
  Smeulders]{uijlings2013selective}
Jasper~RR Uijlings, Koen~EA Van De~Sande, Theo Gevers, and Arnold~WM Smeulders.
\newblock Selective search for object recognition.
\newblock \emph{International journal of computer vision}, 104\penalty0
  (2):\penalty0 154--171, 2013.

\bibitem[Zitnick and Doll{\'a}r(2014)]{zitnick2014edge}
C~Lawrence Zitnick and Piotr Doll{\'a}r.
\newblock Edge boxes: Locating object proposals from edges.
\newblock In \emph{European Conference on Computer Vision}, pages 391--405.
  Springer, 2014.

\bibitem[Mathe et~al.(2016)Mathe, Pirinen, and
  Sminchisescu]{mathe2016reinforcement}
Stefan Mathe, Aleksis Pirinen, and Cristian Sminchisescu.
\newblock Reinforcement learning for visual object detection.
\newblock In \emph{Proceedings of the IEEE Conference on Computer Vision and
  Pattern Recognition}, pages 2894--2902, 2016.

\bibitem[Collobert et~al.(2011)Collobert, Kavukcuoglu, and
  Farabet]{collobert2011torch7}
Ronan Collobert, Koray Kavukcuoglu, and Cl{\'e}ment Farabet.
\newblock Torch7: A matlab-like environment for machine learning.
\newblock In \emph{BigLearn, NIPS Workshop}, number EPFL-CONF-192376, 2011.

\bibitem[Wang et~al.(2016)Wang, Xiong, Wang, Qiao, Lin, Tang, and
  Gool]{Wang2016Temporal}
Limin Wang, Yuanjun Xiong, Zhe Wang, Yu~Qiao, Dahua Lin, Xiaoou Tang, and
  Luc~Van Gool.
\newblock Temporal segment networks: Towards good practices for deep action
  recognition.
\newblock In \emph{European Conference on Computer Vision}, pages 20--36, 2016.

\bibitem[Escorcia et~al.(2016)Escorcia, Heilbron, Niebles, and
  Ghanem]{Escorcia2016DAPs}
Victor Escorcia, Fabian~Caba Heilbron, Juan~Carlos Niebles, and Bernard Ghanem.
\newblock \emph{DAPs: Deep Action Proposals for Action Understanding}.
\newblock Springer International Publishing, 2016.

\bibitem[Heilbron et~al.(2016)Heilbron, Niebles, and Ghanem]{Heilbron2016Fast}
Fabian~Caba Heilbron, Juan~Carlos Niebles, and Bernard Ghanem.
\newblock Fast temporal activity proposals for efficient detection of human
  actions in untrimmed videos.
\newblock In \emph{Computer Vision and Pattern Recognition}, pages 1914--1923,
  2016.

\end{thebibliography}

\end{document}